\pdfoutput=1

\documentclass[11pt]{article}

\usepackage[review]{EMNLP2023}

\usepackage{times}
\usepackage{latexsym}

\usepackage[T1]{fontenc}

\usepackage[utf8]{inputenc}

\usepackage{microtype}

\usepackage{inconsolata}

\usepackage{graphicx}
\usepackage{float}
\usepackage{amsmath} 
\usepackage{multirow} 

\title{IMPROVING FUZZY RULE CLASSIFIER WITH BRAIN STORM OPTIMIZATION AND RULE MODIFICATION}

\author{Yan Huang, Wei Liu, Xiaogang Zang \\
  }

\begin{document}
\maketitle
\begin{abstract}
  The expanding complexity and dimensionality in the search space can adversely affect inductive learning in fuzzy rule classifiers, thus impacting the scalability and accuracy of fuzzy systems. This research specifically addresses the challenge of diabetic classification by employing the Brain Storm Optimization (BSO) algorithm to propose a novel fuzzy system that redefines rule generation for this context. An exponential model is integrated into the standard BSO algorithm to enhance rule derivation, tailored specifically for diabetes-related data. The innovative fuzzy system is then applied to classification tasks involving diabetic datasets, demonstrating a substantial improvement in classification accuracy, as evidenced by our experiments.
 
\end{abstract}

\section{Introduction}

Computational intelligence, such as neural networks, decision trees, and fuzzy logic, is increasingly employed for complex problem-solving \cite{Akbal2018}.Fuzzy rule-based classification systems are particularly valuable in machine learning, offering interpretable solutions in various domains, including security, image analysis, and medical diagnostics \cite{DBLP:conf/acl/ZhuLLX23,DBLP:journals/eswa/ZhuLX24}. However, these classifiers often struggle with high-dimensionality or multiple variables, leading to an exponential increase in the rule search space and impacting system performance and efficiency \cite{tan2012application}.

In our study, we develop a fuzzy logic-based expert system using 'if-then' rules and membership functions to establish a core rule base \cite{varniab2019classification}. 
The interaction between membership functions and rule structures is crucial for the concurrent development of these systems \cite{tsai2018novel}, creating a fuzzy connection between inputs and outputs \cite{shunmugapriya2017hybrid,shi2022cross}. 
Generating these rules from training data in a high-dimensional space presents a significant search challenge within this vast space. 
Gene selection algorithms, such as those employing simplified swarm optimization combined with multi-filter ensemble techniques\cite{lai2021gene}, have demonstrated effective strategies for navigating the complexity of high-dimensional search spaces \cite{tuba2019classification}.

Various methods exist for developing rule-based classifications in fuzzy logic systems, ranging from simple heuristics to complex techniques like fuzzy neural networks, cluster analysis, and evolutionary algorithms \cite{kim2018design,CastroGertrudes2019,liska1994complete}. 
Heuristic approaches inspired by natural phenomena, such as swarm behavior optimization, algorithmic cooling simulations, luminescent insect-based optimization, and bee foraging-inspired techniques, are well-recognized \cite{10.1007/s10489-020-01856-4}\cite{man2017improved,doi:10.1142/S021848851950003X,ilango2019optimization}. Parameter adjustments in these algorithms are crucial for influencing optimization convergence speed \cite{binu2015bfc}.

The Brain Storm Optimization (BSO) algorithm, while effective in fuzzy systems for rule creation, membership function formulation, and knowledge extraction, faces challenges such as low accuracy and slow convergence in scenarios with expanding spatial dimensions or variables \cite{zen2017determining}\cite{sovatzidi2022brainstorming}. This paper addresses these challenges, particularly in large datasets, by introducing a refined BSO approach tailored for such complexities. Our fuzzy classification system is developed in two phases: initially setting membership functions using a uniform distribution, then enhancing BSO with an exponential average approach. This strategy includes constraints on rule length, number, and class distribution, enabling efficient inference rule formulation for data classification, significantly reducing the search space and improving system efficiency.

Our method, utilizing fuzzy rules derived from a uniform distribution of membership functions, improves diabetes detection accuracy in patients compared to the adaptive generalized fuzzy system (AGFS). It exhibits superior sensitivity, specificity, and stability.

\section{Analyzing the Challenges }

The rule base is vital in a fuzzy system, containing critical rules for data categorization. Two key aspects need attention when formulating these rules: rule acquisition and rule optimization strategy. Rule acquisition involves extracting precise rules from large datasets to meet the fuzzy system’s requirements. Rule optimization strategy balances rule length and quantity, aiming for interpretability and accuracy without overburdening computations. Fuzzy systems are preferred in classification for their adaptability and lack of a learning phase, but designing them becomes challenging as the search space grows more complex and dimensional. This article proposes a new approach to address these challenges.

\begin{table*}
\centering
\caption{\label{Table1}
Some rules for the font styles used in figures
\cite{dennis2014agfs,wu2019novel,lakshmi2020improved,MENG201248,you2021ensemble,sarin2023three}
}
\resizebox{\linewidth}{!}{
\begin{tabular}{l l l l}
\hline
Cited Works & Methodology & Objective & Limitations \\
\hline
(2) & Genetic algorithms & Optimization of Rules & Limited applicability
in multi-class scenarios \\
(10) & Ant Colony Optimization & Extraction of Rules & Necessitates Various Objectives in Fuzzy System Design \\
(18)& Algorithm for extracting rules using adaptive genetic approaches & Developing and
refining rules & Simplified member function design approach \\
(19) & Combined Ant and Bee Algorithm Approach & Design of Membership Functions & Inefficacy in Continuous Space Exploration \\
(20) & Algorithm based on genetic grouping & Tendency for global search sensitivity in optimizing rule sets and membership functions & High sensitivity in global search optimization \\
(21) & Genetic algorithm & Extraction of Rules & Optimization requirement for member function designs \\
(22) & Application of Genetic Algorithm & Rule Derivation & Requirement for Rule Weighting \\
(23) & The algorithm is based on genetic grouping & Meta-heuristic algorithms & Randomness can lead to variation in results \\
\hline
\end{tabular}

}

\end{table*}

\section{  Innovations in Data Classification: Brain Storm Enhanced Fuzzy System }
\subsection{Membership Functions Based on Uniform Distribution }

In this subsection, we propose a novel method for developing fuzzy and non-fuzzy membership functions using uniform distribution. The choice of uniform distribution is based on its flexibility and adaptability, allowing for efficient function design without being constrained by specific data distribution positions. The linear-scaled membership mechanism is crucial for assigning variable association degrees to data points, essential for rule formulation. Establishing a membership function involves selecting the function type, determining the number of functions per attribute, and defining each function's boundaries.

The input database can be represented as:

\begin{equation}
\text{D=}\text{X}_{\text{ij}}\text{,0<}\text{i}\text{,}\text{j}\text{<}\text{m}\text{,}\text{n}
\end{equation}

Where   $\text{X}$  denotes the elements in the database,   $\text{m}$  is the total number of attributes, and   $\text{n}$  represents the number of data objects. 

In the rule base, each rule corresponds to a specific attribute cluster, denoted as $\text{A}_{\text{i}}$, where $\text{i}$ identifies and distinguishes each rule.

According to the terminology used in this paper,   $\text{p}$ represents the number of membership functions assigned to each attribute in a rule, indicating the total membership functions per attribute. Each function is defined by three interval parameters:  $\text{a}$  , the lower limit marking the start and affecting the orientation;   $\text{b}$  ,the midpoint, representing the peak with the highest membership degree; and   $\text{c}$  ,the upper limit, concluding the function and influencing its shape and position.

Eq. 2 in our study innovatively illustrates how the attribute vector is used in the membership function to determine the value of   $\text{p}$ .

\begin{equation}
\text{A}_{\text{i}}\text{$\leftrightarrow$ }\text{$\mu$}_{\text{ik}}\text{;0<k<p}
\end{equation}

where $\leftrightarrow$  indicates a correspondence relationship, with $\mu_{ik}$ representing the k-th membership function for the i-th attribute.

In this section, each membership function, characterized by parameters  $\text{a}$ , $\text{b}$  and $\text{c}$  ,divides the attribute vector for$\text{A}_{\text{i}}$  into $\text{p}$   intervals, intervals, each further segmented into smaller intervals. The boundary parameters—minimum, central, and maximum—of these segments are computed using the methodologies described in Eq.3 and Eq.4.

\begin{equation}
\text{$\mu$}_{\text{ik}}\text{=f}\left( {\text{X}_{\text{i}}\text{;a,b,c}} \right)
\end{equation}

\begin{equation}
\text{f}\left( {\text{X}_{\text{i}}\text{;a,b,c}} \right)\text{=}{\text{max}\left( \text{X}_{\text{i}} \right)}\text{$\cdot$}\left( \frac{{\text{max}\left( \text{X}_{\text{i}} \right)}\text{+}{\text{min}\left( \text{X}_{\text{i}} \right)}}{\text{2}} \right)
\end{equation}
   
   Fig.1 presents illustrative examples of ternary membership functions, showcasing two distinct types.

   \begin{figure}
       \centering
       \includegraphics[width=1\linewidth]{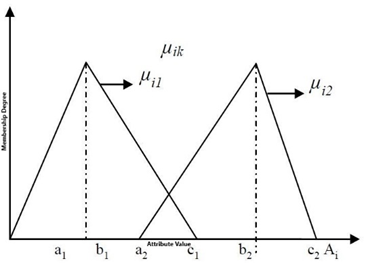}
       \caption{Example of membership functions.}
       \label{fig:1}
   \end{figure}
\subsection{Formulating and Refining Rules via Storm Optimization Technique }
Rule generation and refinement are crucial in fuzzy systems, employing an enhanced BSO algorithm integrated with an Exponential Weighted Moving Average (EWMA) model [18]. This algorithm generates random rules and selects the most effective ones based on rule length, quantity, and class distribution.
\subsubsection{Overall Framework }
\begin{figure}[H] 
    \centering
    \includegraphics[width=1\linewidth]{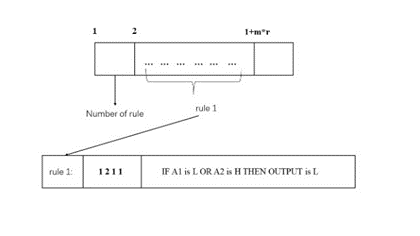}
    \caption{Depiction of Strategy for Addressing Challenges. }
    \label{fig:2}
\end{figure}

Our research introduces an enhanced version of the BSO algorithm tailored for fuzzy system rule structures, incorporating rule sets derived from membership function representations detailed in Eq. 3. This improved algorithm streamlines rule construction by evaluating the number of membership functions per attribute, thus determining the necessary rule count. This process optimizes rule formulation by linking rules with data attributes and output categories, making rule expression more intuitive. The total number of rules,  $\text{r}$   and the size of the program schema,  $\text{1+m*r}$ (where m is the length of individual rules and r the total number),are defined, facilitating optimization and calculation. Fig. 2 illustrates a rule format, with logical operators and output categories specified in its final columns.

\subsubsection{Evaluation Metric for Algorithm Effectiveness }
In every proposed solution, assessing its effectiveness is critical. We introduce a novel fitness metric to appraise solution effectiveness, focusing on optimizing rule number and distribution. This fitness function considers three key factors: rule length, distribution and magnitude, and rule alignment within the database. Methods are designed to balance these factors and, through database input and rule  effectiveness evaluation, improve the rule set.

The fitness function in this paper adopts a maximization approach, focusing on rule length, distribution and magnitude, and alignment within the database. The calculation method is as follows:

\begin{equation}
\text{G=$\alpha$}\text{g}_{\text{1}}\text{+$\beta$}\text{g}_{\text{2}}\text{+$\gamma$}\text{g} _{\text{3}}
\end{equation}

Where  $\alpha$ , $\beta$  and  $\gamma$  serve as weighting constants for the objective functions  $\text{g}_{\text{1}}$ ,  $\text{g}_{\text{2}}$ and  $\text{g} _{\text{3}}$ . The rule length is evaluated by its summation, isolated from rule count and attributes, and normalized to a 0 to 1 scale. If   $\text{g}_{\text{1}}$  close to 1 indicates a short, less complex rule. The primary goal is to maximize this target by minimizing rule length, calculated as follows:

\begin{equation}
\text{g}_{\text{1}}\text{=1-}\left\lbrack \frac{\sum_{\text{i=1}}^{\text{r}}\text{m}_{\text{i}}}{\text{r*m}} \right\rbrack
\end{equation}

where $\text{m}_{\text{i}}$  indicative of each rule's span and r symbolizes the aggregate count of such rules.

Another primary goal is to establish correlations between the rules and the database, assessing by how often each rule appears. A membership function categorizes the database for easy rule comparison. Effective rules are those with frequent matches. To quantify this, we calculate the ratio of the number of rules to total data objects, normalizing in the range of 0 to 1, as follows:

\begin{equation}
\text{g}_{\text{2}}\text{=}\frac{\sum_{\text{i=1}}^{\text{l}}{\text{M}\left( {\text{l}_{\text{i}}\text{,D}} \right)}}{\text{l*n}}
\end{equation}

where  $\text{M}\left( {\text{l}_{\text{i}}\text{,D}} \right)$  signifies the frequency of occurrences of rule  $\text{l}_{\text{i}}$  in correlation with the database, while denotes the overall magnitude of the database.

We aim to stabilize rule variations across categories by analyzing the rule count for each category and calculating their statistical dispersion, normalized to a 0 to 1 scale, as outlined in Eq. 8:

\begin{equation}
\text{g}_{\text{3}}\text{=1-}\frac{\text{V}\left( \text{r,c} \right)}{\text{r}}
\end{equation}

where  $\text{c}$  represents the number of classes and  $\text{V}\left( {\text{r}\text{,}\text{c}} \right)$  stands for the variance function applied to the rule count within each class.

\subsubsection{Process of Brain Storm Optimization Algorithm}
The traditional BSO algorithm has been enhanced by integrating the Exponential Moving Average (EWMA) [18], aimed at improving solution exploration within the search space. This refined algorithm updates solutions each iteration, effectively utilizing global information to enrich evolutionary diversity.

\begin{figure}
    \centering
    \includegraphics[width=0.5\linewidth]{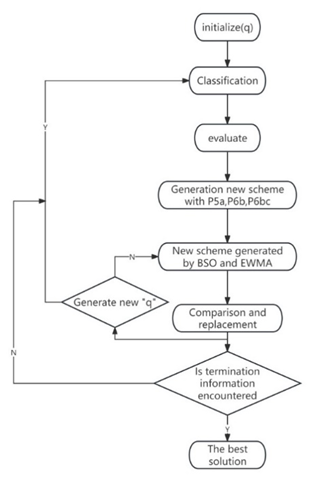}
    \caption{Improved brain storm optimization algorithm flow chart. }
    \label{fig:3}
\end{figure}

In accordance with Fig.3, we elaborate on the operational procedure of the algorithm:

Initialization: We randomly initialize pp strategies in the exploration area    $\text{I}_{\text{i}}\text{=[}\text{I}_{\text{i}\text{1}}\text{,}\text{I}_{\text{i}\text{2}}\text{,}\text{$\cdots$}\text{,}\text{I}_{\text{iq}}\text{]}$  , where i = 1,2…,q, and each "scheme" represents a potential solution.

Classification: We apply membership functions from Eq.2 and Eq.3 to classify data into corresponding classes.

Evaluation: According to Eq. 15, Eq. 16, and Eq. 17 to assess classification results.

Generation of New Scheme: A new scheme is formulated using strategies from Eq. 3 to Eq. 5.

BSO Operation: The BSO algorithm refines the fuzzy system, creating a new configuration.

New scheme generated: Employing the BSO algorithm to enhance the fuzzy system.

Generation of New "q": New candidate solutions $\text{I}_{\text{t}\text{+1}}$ , are generated, adjusted by scaling factor $\vartheta$ and noise term   
 $\text{N}\left( {\text{$\mu$}\text{,}\text{$\sigma$}} \right)$ as per Eq. 12 and Eq. 13.

Comparison and Replacement: New solutions are compared with existing ones using the weighted average method from Eq. 14 for possible replacement.

Is termination information encountered: The process checks if termination criteria are met based on predefined settings.

The Best Solution: The algorithm selects the optimal solution based on the highest performance metrics evaluated.

The candidate schemes are updated based on fitness values. For the schemes detailed in Section 3.2.1, updates incorporate the cluster's central point plus a random variable. In Sections 3.2.2 and 3.2.3, updates use the average central points of clusters. Remaining schemes are adapted using the method in Eq. 9. Specifically, for the schemes in Sections 3.2.2 and 3.2.3, the mean of central points is calculated and used to adjust the scheme to better align with data distribution. Other methods are iteratively updated as outlined in Eq.9, integrating stochastic elements to enhance diversity.

\begin{equation}
\text{I}_{\text{t+1}}\text{=}\text{I}_{\text{t}}\text{+$\xi$N}\left( \text{$\mu$,$\sigma$} \right)
\end{equation}

where  $\text{I}_{\text{t}\text{+1}}$  denotes the random variable at time step $\text{t}\text{+1}$ ,  $\text{t}$  is the time step,  $\xi$ cts as the scaling coefficient for the random variable,   and  $\text{N}\left( {\text{$\mu$}\text{,}\text{$\sigma$}} \right)$  represents a normal distribution with mean  $\mu$  and standard deviation $\sigma$  .

\begin{equation}
\text{$\xi$=S}\text{*}{\text{log}\text{s}}\text{ig}\left( \frac{\frac{\text{N}\text{C}_{\text{max}}}{\text{2-Ne}}}{\text{K}} \right)
\end{equation}

where $\text{s}$   represents a random variable ranging from 0 to 1.  $\text{N}\text{C}_{\text{max}}$  indicates the maximum iteration count, and  $\text{N}\text{C}$  specifies the current iteration. The parameter  $\text{K}$  adjusts the slope of the  ${\text{log}\text{s}}\text{ig}\text{()}$  function. 

\begin{equation}
\text{E}_{\text{t+1}}\text{=$\varepsilon$ }\text{*}\text{I}_{\text{t}}\text{+}\left( \text{1-e} \right)\text{E}_{\text{t}}  
\end{equation}

In this research, we incorporate the Exponential Weighted Moving Average (EWMA) model into our brainstorming-inspired optimization algorithm to improve search efficiency. The mathematical formulation for project forecast using the EWMA model is as follows:

\begin{equation}
\text{I}_{\text{t}}\text{=}\frac{\text{1}}{\text{e}}\left( {\text{E}_{\text{t+1}}\text{-}\left( \text{1-e} \right)\text{*}\text{E}_{\text{t}}} \right)
\end{equation}

Based on Eq.12, we derive Eq.13, which extends the EWMA model by incorporating stochastic perturbations to address uncertainties or noise.

\begin{equation}
   \text{I}_{\text{t+1}}\text{=}\frac{\text{1}}{\text{e}}\left( {\text{E}_{\text{t+1}}\text{-}\left( \text{1-e} \right)\text{*}\text{E}_{\text{t}}} \right)\text{+$\vartheta$N}\left( \text{$\mu$,$\sigma$} \right)
\end{equation}
where  $\vartheta$  i modulates the influence of the normal distribution  $\text{N}\left( {\text{$\mu$}\text{,}\text{$\sigma$}} \right)$  and serves as a tuning parameter to slow down the convergence rate in the evolutionary process.    

\subsubsection{Rule Weight}
In a fuzzy classifier system, assigning weights to rules is crucial for optimizing system performance. With   $\text{r}$   rules, rules, each varying in fuzziness or uncertainty when faced with data, effective allocation of weights is essential. We use two criteria for this: the degree of compliance between rules and the database, and the length of the rules. The formula for assigning weights to rules is:

\begin{equation}
   \text{W}\left( \text{R}_{\text{opt}} \right)\text{=}\frac{\text{1}}{\text{2}}\text{*}\left( {\text{g}_{\text{1}}\text{+}\text{g}_{\text{2}}} \right)
\end{equation}

where  $\text{R}_{\text{opt}}$  is the current rule,  $\text{g}_{\text{1}}$  is the rule length obtained from Eq. 6, and  $\text{g}_{\text{2}}$  is the rule-database match obtained from Eq. 7.

\section{Experiments }
Here is a concise process for data classification in fuzzy systems. After defining membership functions and rules, input data undergoes fuzzification to convert it into fuzzy values based on the membership function definitions. Non-fuzzy computations then calculate a fuzzy evaluation value, representing the degree of membership in specific fuzzy sets, which is used to assign a class label to the input data.

Aligned with earlier concepts, this section evaluates the performance of the fuzzy system introduced in this paper for data classification. The assessment includes experimental  trials and a comprehensive analysis of the results.

\begin{table*}
\centering
\caption{\label{Table2}
Rule Examples for PID Data Set 
}
\resizebox{\linewidth}{!}{
\begin{tabular}{l l l l l l l l l l}
\hline
Rvalue & The plasma glucose concentration in the oral glucose tolerance test is 2 hours(A1) &  Diastolic blood pressure (mm Hg)(A2) & Triceps skin fold \\thickness(mm) (A3) & 2 hours\\ serum insulin\\(mu U/ml) (A4) &  Body mass\\ index (A5) & Age (A6) & Class variable & body weight & And/or method \\
\hline
Rvalue1 & 0 & 0 & 2 & 2 & 2 & 1 & 1 & 0.3333 & 1 \\
Rvalue2 & 2 & 1 & 1 & 1 & 1 & 3 & 2 & 0.5000 & 1 \\
Rvalue3 & 1 & 1 & 0 & 1 & 0 & 1 & 2 & 0.3889 & 1 \\
Rvalue4 & 3 & 1 & 0 & 1 & 0 & 1 & 2 & 0.3333 & 1 \\
Rvalue5 & 1 & 1 & 2 & 2 & 3 & 2 & 2 & 0.4444 & 1 \\
Rvalue6 & 1 & 1 & 3 & 2 & 0 & 2 & 2 & 0.3889 & 1 \\
\hline

\end{tabular}
}
\end{table*}

\subsection{Experimental Settings }
Dataset: The experiment was conducted using the "Pima Indians Diabetes (PID)" dataset from the UCI Machine Learning Repository \cite{meneganti1998fuzzy}. This dataset consists of 768 samples, each described by 8 medical attributes related to the diagnosis of diabetes in Pima Indian women aged 21 and older. The attributes include: (1) number of pregnancies, (2) plasma glucose concentration (after a 2-hour oral glucose tolerance test), (3) diastolic blood pressure (mm Hg), (4) triceps skinfold thickness (mm), (5) 2-hour serum insulin (mu U/ml), (6) body mass index (BMI), (7) diabetes pedigree function, and (8) age. The target variable is binary, indicating whether an individual has been diagnosed with diabetes (1) or not (0). This dataset is widely recognized in the research community for benchmarking classification algorithms due to its moderate complexity and the presence of both continuous and categorical variables. It is commonly used in machine learning tasks to assess the performance of various classifiers, including those related to fuzzy classification systems.

Evaluation Metrics: In comparing the proposed fuzzy system in this paper with a pre-existing system, three key metrics are used: sensitivity, specificity, and accuracy. Below, we define each of these metrics specifically:

\begin{equation}
\text{Sensitivity=}\frac{\text{TP}}{\text{TP+FN}}
\end{equation}

\begin{equation}
\text{Specificity=}\frac{\text{TN}}{\text{TN+FP}}
\end{equation}

\begin{equation}
\text{Accuracy=}\frac{\text{TP+TN}}{\text{TN+FP+TP+FN}}
\end{equation}

"Correct" (TP) indicates correctly identified instances, "False Positive" (FP) represents incorrect identifications, "True Negative" (TN) signifies correct rejections, and "False Negative" (FN) corresponds to incorrect rejections.

The Linux operating system environment with 32 GB of memory was utilized, and initial parameters were set based on empirical knowledge. Using the Brain Storm Optimization algorithm, a set of solutions was generated, and the performance of the current parameter combination was evaluated. Iterative adjustments were made to the values of  $\text{e}$  and  $\text{K}$  based on performance evaluation until the best parameter combination was chosen. The performance of our proposed fuzzy system was compared with the AGFS algorithm, a well-established benchmark in fuzzy rule design known for its adaptive genetic algorithm. This comparison highlighted the advancements our method offers over traditional approaches in fuzzy rule design, with  $\text{e}$  and  $\text{K}$  playing crucial roles in controlling the algorithm's behavior and optimizing its performance.
\subsection{Qualitative Assessment }
Table 2 displays sample rules for the PID dataset, generated using the enhanced Brain Storm Optimization algorithm, crucial for subsequent classification. Our study evaluates the effectiveness of a membership function derived from a uniform distribution in generating and optimizing fuzzy rules, vital in scenarios with large datasets. Traditional approaches do not address the challenge of increased convergence speed when the search space or number of variables expand. Fig. 4 shows that varying   $\text{e}$  and  $\text{K}$  values do not significantly affect classification accuracy, mainly influenced by rule generation. The uniform distribution effectively mitigates position impact, significantly improving fuzzy system accuracy.
\subsection{QUANTITATIVE EVALUATION }
In this section, we quantitatively evaluate the proposed fuzzy system. We start with an in-depth analysis of the system's performance by varying the values of the   $\text{e}$  and  $\text{K}$  constants. Fig.4(1) and 4(2) show how different values of   $\text{e}$  and  $\text{K}$  influence sensitivity, specificity, and accuracy, considering an 80\% proportion of the training set for PID. The results unequivocally demonstrate that the classification performance of the fuzzy system remains stable regardless of variations in $\text{e}$  and  $\text{K}$ . This stability indicates the system's good robustness and adaptability, maintaining high performance under various conditions with strong generalization capabilities.

Convergence speed is crucial for assessing algorithmic performance. This article compares our algorithm's convergence speed with the AGFS algorithm. The x-axis represents the iteration count during training, implying an  increase in the volume of training data. The y-axis measures convergence speed in seconds. The graph demonstrates that as the iteration count—and thus, the volume of training data—increases, our algorithm achieves faster convergence speed.

\begin{figure}
    \centering
    \includegraphics[width=0.5\linewidth]{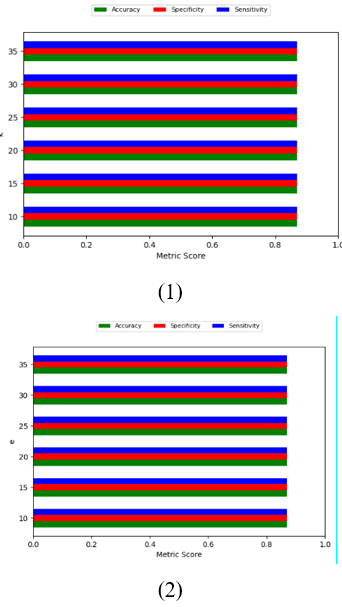}
    \caption{The impact of e and K on the classification performance of the system for PID. }
    \label{fig:4}
\end{figure}

\begin{figure}
    \centering
    \includegraphics[width=1\linewidth]{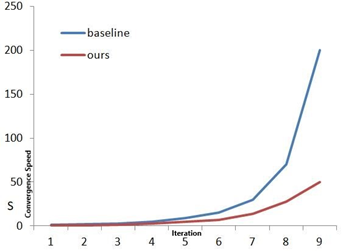}
    \caption{Sample membership function. }
    \label{fig:5}
\end{figure}

\begin{table*}
\centering
\caption{\label{Table3}
Evaluating the Classification Performance of Various Systems on the PID Dataset
}
\resizebox{\linewidth}{!}{
\begin{tabular}{l l l l}
\hline
 & Training-testing data ratio & Benchmark & Enhanced brainstorming 

optimization algorithm \\
\multirow{4}{*}{Sensitive degree} & 0.70 & 0.8651 & 0.8688 \\
 & 0.75 & 0.8658 & 0.8666 \\
 & 0.80 & 0.8621 & 0.8700 \\
 & 0.85 & 0.8548 & 0.8624 \\
\multirow{4}{*}{Special different

Sex} & 0.70 & 0.8651 & 0.8688 \\
 & 0.75 & 0.8658 & 0.8666 \\
 & 0.8 & 0.8621 & 0.8700 \\
 & 0.85 & 0.8548 & 0.8624 \\
\multirow{4}{*}{Quasi-ndeed Sex} & 0.7 & 0.8651 & 0.8688 \\
 & 0.75 & 0.8658 & 0.8666 \\
 & 0.8 & 0.8621 & 0.8700 \\
 & 0.85 & 0.8548 & 0.8624 \\
\hline
\end{tabular}
}
\end{table*}

The algorithm's convergence speed is a crucial benchmark for assessing its quality in practical applications. We conducted experiments to compare the convergence speed of the improved algorithm with the original Brain Storm algorithm. The results in Fig. 5 clearly demonstrate that the new algorithm achieves faster convergence, particularly with increasing volumes of training data.

This article addresses challenges in classification accuracy and convergence speed in high-dimensional data spaces, comparing our method with the AGFS algorithm. By strategically designing membership functions and applying constraints, we significantly reduce the search space, enhancing convergence speed in data classification. Our approach demonstrates superiority over AGFS. Experimental results show that our method significantly enhances classification convergence speed. When data volume sharply increases, optimal rule selection is influenced by rule length, number, and class distribution, partially mitigating the issue of increasing rules. The AGFS algorithm focuses on enhancing fuzzy system classification accuracy, but it's essential to evaluate our optimization method's impact on accuracy. Our improved algorithm offers advantages such as better global information utilization and enhanced evolutionary diversity, aiming to improve both classification accuracy and convergence speed.

Table 3 demonstrates that our approach achieves accuracy levels equivalent to or better than AGFS, especially with a notable increase in the proportion of training data. 
The table highlights improvements in sensitivity, specificity, and accuracy metrics for our method compared to AGFS, particularly with the PID dataset. 
To ensure the experiment's applicability in various scenarios, we use a randomized selection method for choosing training data proportions, including \{70\%, 75\%, 80\%, 85\% \}. 
The enhanced Brain Storm optimization algorithm consistently outperforms conventional methods, with statistically significant improvements. For example, with only 70\% of the data for training, our method achieves an accuracy rate of 86.88\%, compared to AGFS at 86.51\%. With a 75:25 training-testing data ratio, our method and AGFS achieve accuracy rates of 86.58\% and 86.66\%, respectively. With an 80\% training data size, our method reaches its peak performance at 87.00\%, surpassing AGFS at approximately 86.21\%. These results consistently demonstrate the superior effectiveness of our improved Brain Storm algorithm across various training data scenarios compared to AGFS.
\section{CONCLUSION }
To tackle the challenge of diminished scalability and accuracy in fuzzy systems when confronted with the rapid expansion of dimensionality or variables, this paper introduces a modified approach to the conventional rule definition process specifically for diabetic classification. It presents a groundbreaking method for generating fuzzy system rules through an enhanced Brain Storm optimization algorithm, incorporating an exponential model to better address the nuances of diabetes data. For membership function design, a straightforward technique based on uniform distribution is adopted. The fuzzy scores generated by this system are utilized for the classification of diabetic samples, demonstrating superior performance compared to the existing AGFS approach. Future research will aim to further enhance the system’s impact on diabetic classification by replacing the distribution-based method with a specifically designed optimization procedure for the membership function.
\section{ACKNOWLEDGMENT}
This work was supported by the Foundation and Cutting-Edge Technologics Rescarch Programof Ilcnan Province (232102210107), the National Natural Scicnce Foundation of China (grant no.61702516), and the Open Fund of Scicncc and Technology on Thcrmal Energy and Power Laboratory(No. TPL2020C02). Wuhan 2nd Ship Design and Rescarch Institutc, Wuhan, P.R. China.

\bibliography{anthology,custom}
\bibliographystyle{acl_natbib}

\appendix

\end{document}